\documentclass[12pt]{article}

\usepackage[a4paper,margin=1in]{geometry}
\usepackage[T1]{fontenc}
\usepackage{tgtermes}
\usepackage{setspace}
\usepackage{graphicx}
\usepackage{caption}
\usepackage{placeins}
\usepackage{booktabs}
\usepackage{multirow}
\usepackage{makecell}
\usepackage{textcomp}
\usepackage{url}
\usepackage[hidelinks]{hyperref}
\usepackage[numbers,square,sort&compress]{natbib}
\newcommand{\EndoGazeBibliography}{%
  \bibliographystyle{unsrtnat}%
  \bibliography{references}%
}

\graphicspath{{figure/}}
\title{Gaze responses to false-positive computer-aided detection prompts during colonoscopy: a paired-video and real-time eye-tracking study}
\author{%
\parbox{0.95\textwidth}{\centering
Te Luo\textsuperscript{1,2,\textdagger}, Yan Zhu\textsuperscript{3,4,\textdagger},
Peiyao Fu\textsuperscript{3,4}, Ruijie Yang\textsuperscript{1,2,5,6},\\
Xian Yang\textsuperscript{7}, Quanlin Li\textsuperscript{3,4,*},
Pinghong Zhou\textsuperscript{3,4,*}, Shuo Wang\textsuperscript{1,2,*}\\[0.75em]
\small \textsuperscript{1}Digital Medical Research Center, School of Basic Medical Sciences, Fudan University, Shanghai, China\\
\small \textsuperscript{2}Shanghai Key Laboratory of MICCAI, Shanghai, China\\
\small \textsuperscript{3}Endoscopy Center and Endoscopy Research Institute, Zhongshan Hospital, Fudan University, Shanghai, China\\
\small \textsuperscript{4}Shanghai Collaborative Innovation Center of Endoscopy, Shanghai, China\\
\small \textsuperscript{5}Zhejiang University, Hangzhou, China\\
\small \textsuperscript{6}Shanghai Institute for Advanced Study, Zhejiang University, Shanghai, China\\
\small \textsuperscript{7}Alliance Manchester Business School, The University of Manchester, Manchester, United Kingdom\\[0.5em]
\small \textsuperscript{\textdagger}These authors contributed equally to this work.\\
\small \textsuperscript{*}Corresponding authors: Quanlin Li <\href{mailto:li.quanlin@zs-hospital.sh.cn}{li.quanlin@zs-hospital.sh.cn}>;
Pinghong Zhou <\href{mailto:zhou.pinghong@zs-hospital.sh.cn}{zhou.pinghong@zs-hospital.sh.cn}>; Shuo Wang <\href{mailto:shuowang@fudan.edu.cn}{shuowang@fudan.edu.cn}>.}}
\date{}

\begin{document}
\maketitle
\doublespacing

\newpage

\section*{Abstract}

\subsection*{Background and Aims}

Computer-aided detection (CADe) systems use bounding-box prompts to direct endoscopists' attention to suspected lesions during colonoscopy. However, CADe also generates false-positive prompts, which may divert visual attention during lesion search. Although their frequency has been widely reported, the attentional impact of individual false-positive prompts remains unclear. We used event-locked eye tracking to quantify gaze attraction and attention occupation following false-positive CADe prompts in controlled paired-video and prospective real-time clinical settings.

\subsection*{Methods}

We conducted complementary retrospective and prospective eye-tracking studies. In the retrospective paired-video experiment, 3 senior and 2 novice endoscopists viewed 60 prerecorded colonoscopy videos under both unassisted and CADe-assisted conditions; in the prospective study, gaze was recorded during 42 real-time CADe-assisted colonoscopies performed by 9 senior endoscopists. Expert-annotated lesion windows defined lesion events, while screened CADe prompts outside these windows were classified as false-positive artifact events. Event-locked analyses quantified gaze attraction, attention occupation, recovery, and time amplification following artifact events; lesion ROI recognition and first-entry time were assessed as secondary outcomes. The retrospective experiment enabled within-subject comparison under controlled conditions, whereas the prospective recordings assessed whether the same gaze responses were observed during real-time clinical workflow.

\subsection*{Results}

False-positive CADe prompts attracted gaze in 48.6\% (68/140) of retrospective observations and 65.2\% (533/817) of prospective events. Among attraction events with complete recovery, median attention occupation lasted 1000 ms in the retrospective dataset and 1100 ms in the prospective dataset, substantially longer than the corresponding median prompt durations of 33 ms and 267 ms. This yielded median time amplifications of 17.55-fold and 5.15-fold, respectively. In the retrospective paired analysis, visible artifact prompts also shifted gaze closer to the prompted region than the same-coordinate unassisted reference. As a secondary analysis, lesion gaze recognition was high without and with CADe assistance (98.0\% vs 99.0\%), while first gaze entry into lesion ROIs occurred 147.8 ms earlier with CADe assistance.

\subsection*{Conclusions}

False-positive CADe prompts frequently captured endoscopists' gaze, with attention persisting beyond prompt visibility and showing substantial temporal amplification. Prompt-related attentional burden should be considered in future CADe evaluation and design to support more effective human--AI collaboration.

\newpage

\section{Introduction}\label{sec:introduction}

Artificial intelligence (AI)-driven computer-aided detection (CADe) systems are increasingly used in real time during colonoscopy to support lesion detection. Randomized controlled trials and meta-analyses have shown that CADe assistance can improve the adenoma detection rate (ADR) and adenomas per colonoscopy (APC) and reduce the adenoma miss rate (AMR) \citep{wang_real_time_polyp_detection,hassan_real_time_cade_meta_analysis}. However, an improvement in ADR has not been consistently observed in pragmatic implementation and population-based screening trials \citep{ladabaum_pragmatic_cade_trial,davila_pinon_computer_assisted_colonoscopy}. Beyond detection outcomes, the clinical value of CADe also depends on how its use affects the behavior and performance of the endoscopist during human--AI collaboration. Recent studies have raised, but not resolved, concerns about whether repeated CADe exposure alters unassisted detection ability. Budzyń et al.~\citep{budzyn_endoscopist_deskilling} reported a lower unassisted ADR after a period of AI exposure, whereas Pedersen et al.~\citep{pedersen_learning_deskilling} found no persistent upskilling or deskilling after CADe withdrawal. Eye-tracking studies have also shown that CADe can alter visual-search behavior and influence how endoscopists respond to AI-marked regions \citep{troya_cade_reaction_time_gaze,ishibashi_ai_site_recognition,ito_cade_gaze_shift,zhu_endoscopist_ai_gaze_dataset}. Together, these findings highlight human--AI interaction as an important dimension of CADe evaluation, in line with reporting guidance for the early-stage clinical evaluation of AI-based decision support \citep{vasey_decide_ai}. Beyond whether AI improves detection, it is necessary to understand how its prompts shape the endoscopist's attention during visual search.

The intended function of a CADe prompt is to direct the endoscopist's attention toward a candidate lesion. However, the same interface also generates recurrent false-positive prompts over non-lesion regions. Abruptly appearing, task-irrelevant visual objects can capture gaze during goal-directed visual search \citep{theeuwes_oculomotor_capture}, so false-positive prompts may compete with the endoscopist's own visual search. Published studies have reported tens of false-positive activations per colonoscopy, often arising from bowel-wall artifacts \citep{hassan_false_positive_classification}, and endoscopists may spend additional procedural time attending to them \citep{okumura_false_positive_detection}. Reported false-positive rates vary according to how an alert is defined \citep{holzwanger_false_positive_definitions}, and a greater false-positive burden may reduce the detection benefit provided by CADe \citep{zhang_false_positive_assistance}. Nevertheless, existing evaluations have primarily quantified false-positive prompts by their frequency or procedure-level burden. A controlled video study also examined clinical assessments and decision times following true- and false-positive AI recommendations \citep{vanberkel_false_positive_continuous_interaction}. These approaches characterize alert burden and decision-making, but do not resolve whether an individual prompt captures gaze, how long it occupies visual attention, or whether its attentional effect persists after the prompt disappears. Eye tracking provides a direct means of examining this prompt-level visual response. Previous studies have characterized experience-related visual-search patterns \citep{almansa_visual_gaze_adenoma,he_simulated_colonoscopy_gaze}, examined associations between gaze allocation and polyp and adenoma detection \citep{lami_gaze_patterns_polyp_detection,nagai_optimal_visual_gaze}, and assessed gaze responses to CADe-marked regions \citep{troya_cade_reaction_time_gaze,ishibashi_ai_site_recognition,ito_cade_gaze_shift,zhu_endoscopist_ai_gaze_dataset}. However, how an individual false-positive CADe prompt attracts, occupies, and subsequently releases the endoscopist's gaze remains insufficiently characterized.

To address this gap, we shifted the unit of analysis from procedure-level outcomes to individual CADe prompts and used event-locked eye tracking to characterize the temporal course of prompt-related visual attention. We quantified gaze attraction, attention occupation, recovery, and the persistence of attention relative to prompt visibility following false-positive CADe prompts. We combined a controlled paired-video experiment with prospective real-time clinical recordings to examine these responses under both experimental and clinical conditions, while lesion-directed gaze responses were evaluated as a secondary comparator. This event-level approach provides a process-level assessment of human--AI interaction that complements conventional detection endpoints and false-positive event counts.

\section{Methods}\label{sec:methods}

\subsection{Study design}\label{subsec:study-design}
 
\begin{figure}[!htbp]
\centering
\includegraphics[width=0.85\textwidth]{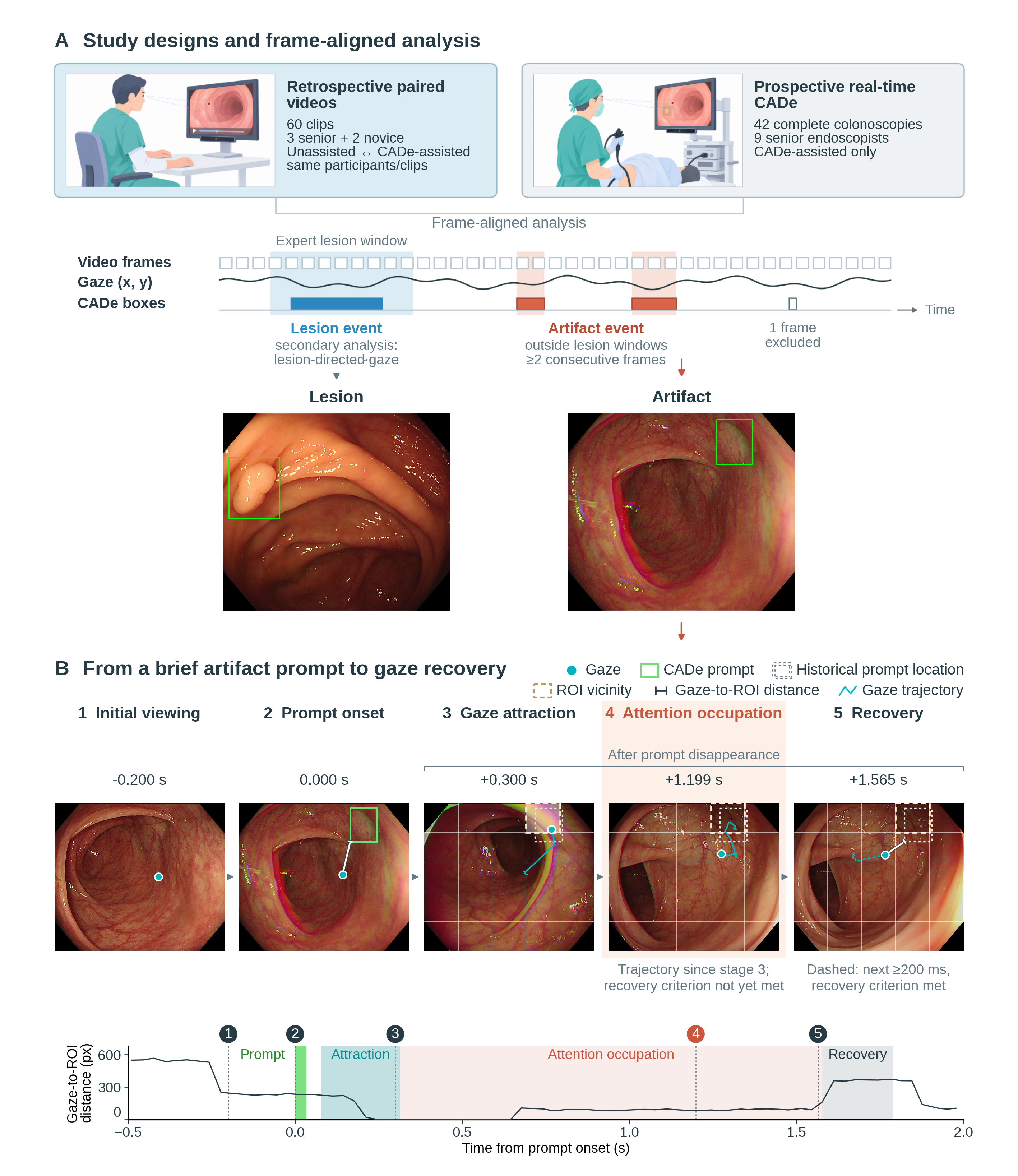}
\caption{\textbf{Study designs and frame-aligned analysis of prompt-related gaze responses.}
\textbf{(A)} The retrospective experiment paired unassisted and CADe-assisted viewing of the same 60 clips by 3 senior and 2 novice endoscopists. The prospective dataset comprised 42 complete CADe-assisted colonoscopies performed by 9 senior endoscopists, without an unassisted control. In both datasets, video frames, gaze coordinates, and CADe bounding boxes were aligned to a common timeline. Artifact events were defined as CADe boxes outside expert-annotated lesion windows that persisted for at least two consecutive frames; single-frame detections were excluded. Lesion-directed gaze within lesion windows was assessed as a secondary analysis. Example frames show CADe prompts over a lesion and a non-lesion region.
\textbf{(B)} Five frames from one retrospective CADe-assisted recording illustrate initial viewing, prompt onset, gaze attraction, attention occupation after prompt disappearance, and recovery. Turquoise markers indicate gaze position, and green boxes indicate visible CADe prompts. White dashed rectangles mark the former prompt location. Pale-yellow dashed grid cells define the region-of-interest (ROI) vicinity on a screen-fixed $5\times5$ grid, which does not track tissue motion. White capped lines indicate the shortest distance from gaze position to the former prompt box. Turquoise lines connect successive valid gaze samples, from prompt onset to the current frame in stage 3 and from the stage-3 sample to the current frame in stage 4. At stage 4, gaze has left the ROI vicinity but has not yet met the recovery criterion. Stage 5 marks the start of a subsequently confirmed recovery interval, with stable non-attraction outside the ROI vicinity for at least 200 ms; the dashed line traces gaze through this interval. Frame times are relative to prompt onset; equal spacing does not indicate equal temporal intervals. The lower strip plots the shortest gaze-to-prompt-box distance in pixels from 0.5 s before to 2 s after prompt onset. The green bar marks the visible prompt, and shaded bands mark the attraction segment, the attention-occupation interval, and the recovery interval. Numbered dashed lines mark the five frames.}
\label{fig:workflow}
\end{figure}

We conducted an eye-tracking study comprising a retrospective paired-video experiment and prospective real-time CADe recordings (Figure~\ref{fig:workflow}A). 
In the retrospective experiment, endoscopists viewed matched prerecorded colonoscopy videos under unassisted and CADe-assisted conditions, enabling within-subject comparisons of lesion-directed gaze and gaze responses to false-positive prompts. In the prospective component, gaze was recorded during live CADe-assisted colonoscopies to assess whether the same prompt-related gaze responses were observed during routine clinical workflow. Because the two datasets differed in study design, control condition, and event acquisition, they were analyzed and reported separately. Data for both components were drawn from the eye-tracking dataset previously published by Zhu et al., who described the acquisition procedures in detail \citep{zhu_endoscopist_ai_gaze_dataset}. The study protocol was approved by the Institutional Review Board of Zhongshan Hospital (Approval No.~B2023-262R). All participating endoscopists and patients provided written informed consent for study participation and data sharing. Procedure-related clinical information was collected without personally identifying information.

\FloatBarrier

\subsection{Subjects}\label{subsec:subjects}

The retrospective paired-video experiment included 5 endoscopists, comprising 3 senior and 2 novice endoscopists, each of whom completed both the unassisted and CADe-assisted viewing sessions. The prospective dataset included 42 complete real-time CADe-assisted colonoscopies performed by 9 senior endoscopists. Senior endoscopists were defined as those with experience of at least 5000 colonoscopy procedures. Two senior endoscopists (Y.Z. and P.Y.F.) independently reviewed the videos for lesion annotation across both datasets.

\subsection{Eye-tracking setup and CADe system}\label{subsec:eye-tracking-apparatus-and-cade-prompt-source}

Gaze was recorded using a Tobii Pro Nano eye tracker (Tobii Technology, Stockholm, Sweden) at a sampling rate of 30 Hz. In the retrospective experiment, the eye tracker was calibrated for each participant before recording. Videos were displayed on a 15.6-inch Full HD monitor (1920 $\times$ 1080 pixels), with participants seated approximately 65--70 cm from the screen. In the prospective study, the eye tracker was mounted below the endoscopy monitor and calibrated using a standard 9-point procedure before each colonoscopy. Calibration accuracy was verified using a test fixation point, and recalibration was performed when the deviation exceeded 0.5 degrees. Gaze coordinates and validity information for both eyes were recorded continuously and synchronized with the corresponding video frames using the native frame rate of each recording.

The CADe system was EndoAdd (June 2023 version; Xuan Wei Technology, China), which displayed real-time bounding-box prompts over regions identified as suspected lesions. These visible bounding boxes were used to define lesion and false-positive artifact regions of interest (ROIs) for subsequent gaze analysis. Video frames, CADe bounding boxes, and gaze data were temporally aligned before event-level analysis.

\subsection{Expert lesion annotation}\label{subsec:expert-lesion-annotation}

Two expert endoscopists independently reviewed all retrospective and prospective videos to identify lesions and annotate their on-screen appearance and disappearance times. Disagreements were resolved by consensus, and the resulting intervals defined the lesion windows used throughout the analysis. In the prospective recordings, cecal intubation time was additionally annotated to define the start of withdrawal. Lesion-window onset was used as (t=0) for all lesion-directed gaze analyses.

\subsection{Retrospective paired-video experiment}\label{subsec:retrospective-paired-video-experiment}

All participants viewed the same set of 60 prerecorded colonoscopy clips, each approximately 30 s in duration, concatenated into a single sequence with a total duration of 33 min 23 s. Two matched versions of the sequence were prepared: an unassisted version without CADe prompts and a CADe-assisted version with visible bounding-box prompts. Each participant completed both viewing sessions, with the unassisted session performed first and the CADe-assisted session two weeks later. Screen gaze was recorded throughout both sessions. 

Video frames, per-frame CADe bounding boxes, and expert-annotated lesion windows were aligned to a common timeline and combined with the participant-specific gaze recordings (Figure~\ref{fig:workflow}A). For lesion analysis, the visible CADe bounding box defined the lesion region of interest (ROI) in the CADe-assisted condition. The same spatial coordinates at the corresponding time points were used as the reference ROI in the unassisted condition, enabling within-subject comparison of gaze responses to the same lesion region.

For false-positive prompt analysis, CADe bounding boxes occurring outside expert-annotated lesion windows were identified as candidate artifact events and screened before analysis. Events were required to persist for at least two consecutive detection frames. The final dataset comprised 41 lesion events and 28 artifact events, each evaluated across both viewing conditions.

\subsection{Prospective real-time CADe recordings}\label{subsec:prospective-real-time-cade-recordings}

The prospective dataset comprised real-time eye-tracking recordings obtained during CADe-assisted colonoscopy. For each procedure, the endoscopy video, gaze coordinates, and per-frame CADe bounding boxes were synchronized to a common timeline. Cecal intubation marked the start of withdrawal, and expert-annotated lesion appearance and disappearance times defined the lesion windows.

CADe bounding boxes overlapping a lesion window were assigned to the corresponding lesion ROI, whereas boxes occurring outside all lesion windows were identified as candidate false-positive artifact events. After event screening, the final dataset comprised 60 lesion windows and 817 artifact events from 42 complete colonoscopies. Lesion-directed gaze and false-positive prompt-related gaze responses were analyzed separately as descriptive measures of real-time human--AI interaction during clinical CADe use.

\subsection{Gaze metrics}\label{subsec:gaze-outcome-metrics}

We used event-locked gaze analysis to characterize attraction, attention occupation, recovery, and time amplification following false-positive CADe prompts (Figure~\ref{fig:workflow}B). Lesion-directed gaze recognition and first ROI-entry time were assessed as secondary measures within expert-annotated lesion windows (Figure~\ref{fig:workflow}A).

Lesion-directed gaze metrics quantified whether and how rapidly the endoscopist's gaze entered the lesion region of interest (ROI) or its matched reference region. Lesion gaze recognition was defined as cumulative gaze within the ROI for $\geq100$ ms during the expert-annotated lesion window. First ROI-entry time was defined as the interval from lesion-window onset ($t=0$) to the first gaze sample within the ROI. In the retrospective experiment, the visible CADe bounding box defined the lesion ROI in the CADe-assisted condition, while the same spatial coordinates at the corresponding time points defined the reference ROI in the unassisted condition.

False-positive prompt-related gaze metrics quantified gaze attraction, attention occupation, recovery, and temporal amplification following artifact onset. Artifact event rate was calculated as the number of artifact events per minute of withdrawal time. Each artifact event was analyzed relative to its onset ($t=0$) using an event-locked window.

For visualization of the event-locked gaze response, this interval was divided into 0.5-s bins. Within each bin, the ROI gaze-hit proportion was calculated as the proportion of valid gaze samples falling within the event-specific artifact ROI. The artifact ROI was defined as the union of bounding boxes observed during the first 0.5 s after artifact onset and was held fixed throughout the event-locked window.

Gaze attraction was assessed during the first 0.5 s after artifact onset using the spatial distance between gaze position and the artifact ROI. Distance was defined as zero when gaze fell within the ROI and otherwise as the shortest pixel distance to the ROI. For events containing multiple boxes, the minimum distance to any box was used. A gaze sample was considered direction-gated when its distance to the artifact ROI was smaller than that measured 100 ms earlier. An event was classified as attraction-triggered when the 0--0.5 s post-onset window contained at least 300 ms of valid gaze and a contiguous direction-gated approach lasting $\geq100$ ms.

Immediate attraction magnitude $A_{\mathrm{peak}}$ was defined as the difference between the median baseline gaze-to-ROI distance during the -0.5 to 0 s interval and the minimum direction-gated distance during the first 0.5 s after prompt onset. The time at which this minimum distance occurred was denoted $t_{\mathrm{peak}}$.

Attention occupation duration was operationally defined as the interval from the onset of the qualifying direction-gated approach to recovery. This onset was the time of the first gaze sample in the first contiguous direction-gated approach lasting $\geq100$ ms within 0--0.5 s after prompt onset. The search for recovery began at $t_{\mathrm{peak}}$ for each event. Recovery was defined as stable non-attraction lasting $\geq200$ ms. For events in which gaze had entered the ROI vicinity, recovery additionally required gaze to move outside that vicinity. ROI vicinity was defined using the display-wide $5\times5$ grid cells containing the centers of artifact bounding boxes observed during the first 0.5 s after onset. Events without recovery within the 5-s observation window were treated as right-censored.

Time amplification was calculated for recovered attraction events as the ratio of attention occupation duration to artifact on-screen duration. Artifact on-screen duration was defined as the interval between artifact onset and disappearance. Attention occupation duration was treated as the primary temporal measure of prompt-related gaze persistence, with time amplification used as a secondary duration-normalized measure. The same event-locked framework was applied separately to the retrospective and prospective datasets.

\subsection{Statistical analysis}\label{subsec:statistical-analysis}

All analyses were descriptive, and no formal hypothesis tests were performed. The study was designed to characterize prompt-related gaze mechanisms rather than to test a prespecified hypothesis, and the retrospective experiment included only five endoscopists whose observations were nested within subjects and within lesion or artifact events. We therefore report point estimates, sample sizes, and interval estimates rather than P values. The retrospective and prospective datasets were analyzed and reported separately and were not pooled.

Binary outcomes, including lesion gaze recognition and attraction triggering, are reported as proportions with numerators and denominators. Ninety-five percent confidence intervals for attraction-trigger rates were calculated with the Wilson score method. Continuous outcomes are reported as participant-level means for retrospective first ROI-entry time and as medians for right-skewed duration and ratio measures, namely artifact on-screen duration, attention occupation duration, and time amplification. Event-locked artifact-ROI gaze hit rates are shown as the mean across events within each 0.5-s bin, with 95\% confidence intervals computed as the mean $\pm$ 1.96 standard errors. These intervals treat observations as independent and do not adjust for clustering within endoscopists or procedures; they are intended as descriptive indicators of precision.

Because each retrospective endoscopist viewed the same events under both conditions, comparisons between unassisted and CADe-assisted viewing were within-subject. For retrospective first ROI-entry time, values were first averaged within each participant and then summarized across the five paired participants. Immediate attraction was compared as the mean paired difference in direction-gated gaze-to-ROI distance over the complete subject--artifact observations evaluable in both conditions, with equal weight per observation. Prospective first ROI-entry time was summarized as the median at the lesion-window level among recognition-positive windows. Attention occupation duration and time amplification were summarized among attraction-triggered events with observed recovery; events without recovery within the 5-s window were right-censored, excluded from these summaries, and counted separately. Experience-stratified results for senior and novice endoscopists were exploratory. Differences were computed from unrounded values, so a reported difference may deviate in the last digit from the subtraction of the rounded values shown. All analyses were performed in Python (version 3.13) using NumPy (version 2.4), pandas (version 3.0), and Matplotlib (version 3.10).

\section{Results}\label{sec:results}

\subsection{Study datasets and event characteristics}\label{subsec:study-datasets-and-cade-context}

The retrospective paired-video dataset included 41 lesion events and 28 false-positive artifact events per condition, each evaluated by 5 endoscopists (3 senior and 2 novice), yielding 205 subject--lesion and 140 subject--artifact observations per condition. The prospective dataset included 60 lesion events and 817 false-positive artifact events from 42 real-time CADe-assisted colonoscopies performed by 9 senior endoscopists. The false-positive event rate was 0.84 events/min in the retrospective dataset (28 events over 33.4 min) and 2.77 events/min in the prospective dataset (817 events over 294.9 min) (Table~\ref{tab:cade-context}).

\begin{table}[!htbp]
\centering
\singlespacing
\footnotesize
\setlength{\tabcolsep}{5pt}
\renewcommand{\arraystretch}{1.15}
\caption{Comparison of lesion detection and false-positive event rates across CADe systems}
\label{tab:cade-context}
\newsavebox{\cadetabbox}
\sbox{\cadetabbox}{%
\begin{tabular}{@{}lllcc@{}}
\toprule
CADe system & Source & Setting & Detection sensitivity & \makecell[c]{False-positive\\events (/min)} \\
\midrule
EndoAdd & Present study & Retrospective & 41/41 lesion windows (100\%) & 0.84 \\
EndoAdd & Present study & Prospective & 60/60 lesion windows (100\%) & 2.77 \\
\addlinespace[3pt]
GI Genius & \citep{hassan_ai_validation} & Retrospective validation & 99.7\% (337/338) & --- \\
GI Genius & \citep{hassan_false_positive_classification} & Post-hoc RCT videos & --- & 2.4 $\pm$ 1.2 \\
A SYSTEM & \citep{chung_cade_false_positive_comparison} & Prospective & 100\% & 3.28 \\
B SYSTEM & \citep{chung_cade_false_positive_comparison} & Prospective & 100\% & 1.24 \\
Deep-GI & \citep{tiankanon_false_positive_thresholds} & Prospective & 99.4\% & 1.55 \\
CAD EYE (Fujifilm) & \citep{tiankanon_false_positive_thresholds} & Prospective & 85.0\% & 2.87 \\
\bottomrule
\end{tabular}}
\usebox{\cadetabbox}\par
\begin{minipage}{\wd\cadetabbox}
\scriptsize
\textit{Notes:} The present study used EndoAdd (June 2023 version; Xuan Wei Technology, China). Lesion coverage for EndoAdd denotes visible CADe prompts within expert-annotated lesion windows, whereas published studies report per-lesion detection sensitivity. False-positive event rates were calculated over withdrawal time. Values for Deep-GI and CAD EYE were derived from reported alert counts and withdrawal times at the $\geq$0.5-s threshold. Published values are provided for context and do not represent direct head-to-head comparisons. ---, not reported.
\end{minipage}
\end{table}
\FloatBarrier

\subsection{Gaze responses to false-positive CADe prompts}\label{subsec:artifact-roi-gaze-attraction-and-attention-occupation}

\begin{figure}[!htbp]
\centering
\includegraphics[width=\textwidth]{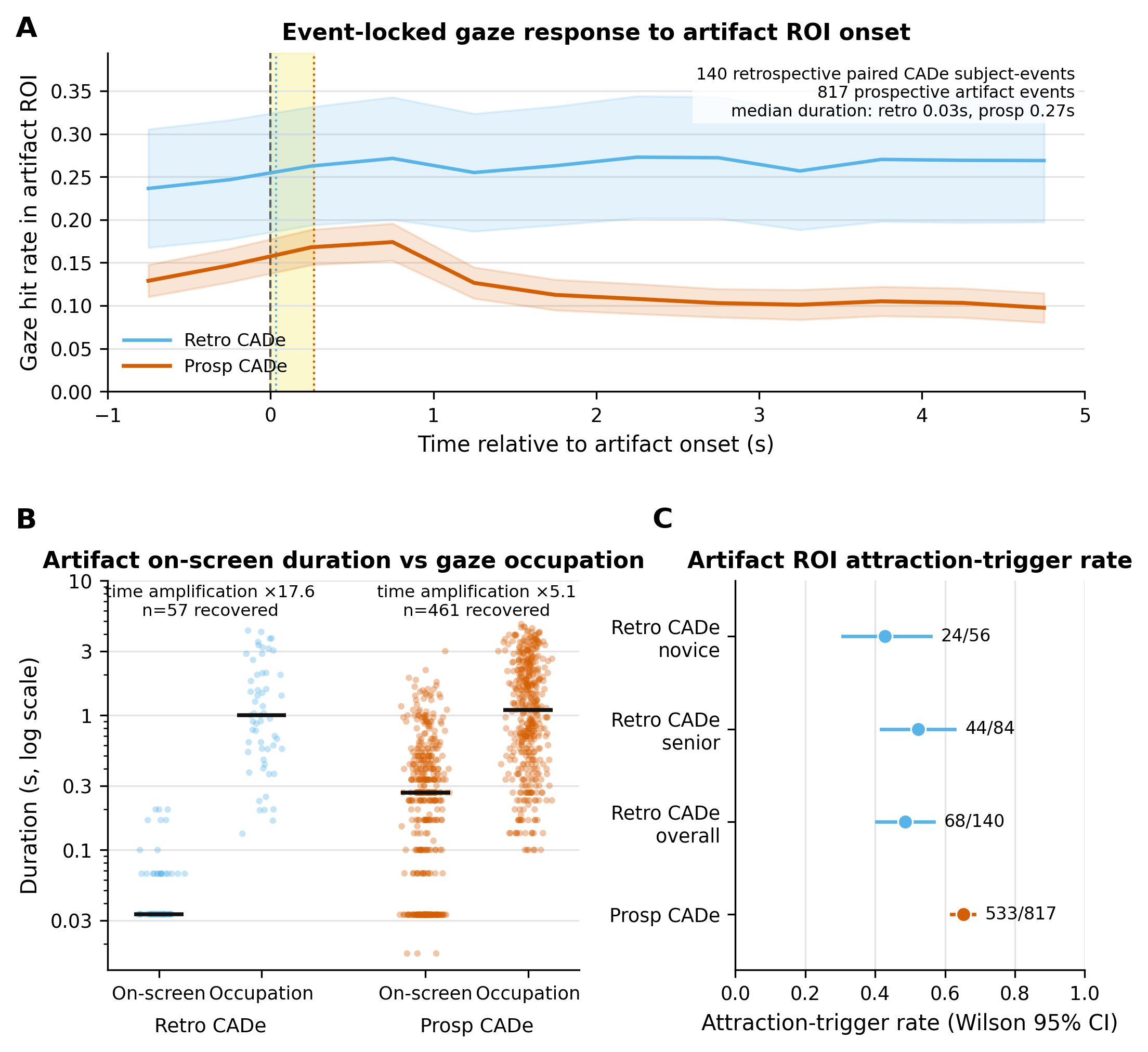}
\caption{Artifact ROIs captured gaze and produced attention occupation. (A) Event-locked gaze response to artifact ROI onset in retrospective paired CADe subject-events and prospective real-time CADe artifact events. Lines show the mean artifact-ROI gaze hit rate across 0.5-s time bins (1.0 s before to 5.0 s after onset), shaded bands show 95\% confidence intervals, time zero marks artifact onset, and vertical reference lines mark median artifact on-screen duration in the retrospective and prospective datasets. This panel provides descriptive trajectory context and the datasets are not pooled. (B) Artifact on-screen duration and gaze attention occupation duration among recovered attraction events, shown on a log-scaled duration axis for retrospective and prospective CADe data. Median bars summarize the distributions, and the annotated time-amplification values quantify how brief artifact prompts corresponded to longer gaze occupation. Right-censored events were excluded from this recovered-event panel as a conservative complete-recovery subset. (C) Artifact ROI attraction-trigger rates with Wilson 95\% confidence intervals for retrospective novice, retrospective senior, retrospective overall, and prospective CADe events. Retrospective paired data provide the controlled artifact-gaze evidence, while prospective real-time CADe data show that the same artifact-gaze mechanism was observable during clinical workflow.}
\label{fig:artifact}
\end{figure}

False-positive prompts attracted gaze in 48.6\% (68/140; Wilson 95\% CI, 40.4--56.8\%) of retrospective observations and 65.2\% (533/817; 95\% CI, 61.9--68.4\%) of prospective events (Figure~\ref{fig:artifact}C). In the retrospective paired analysis, direction-gated gaze approach toward the prompted region was greater with visible CADe prompts than with the same-coordinate unassisted reference (mean paired difference, +25.7 px; n=121).

Among attraction-triggered events with complete recovery, median attention occupation duration was 1000 ms in the retrospective dataset (n=57) and 1100 ms in the prospective dataset (n=461), substantially longer than the corresponding median prompt durations of 33 ms and 267 ms. Median time amplification was 17.55-fold and 5.15-fold, respectively (Figure~\ref{fig:artifact}B). These estimates were based on 57 of 68 retrospective and 461 of 533 prospective attraction-triggered events with observed recovery; the remaining events were right-censored at the end of the observation window (Table~\ref{tab:results}).

\FloatBarrier

\begin{table}[!htbp]
\centering
\singlespacing
\footnotesize
\setlength{\tabcolsep}{8pt}
\renewcommand{\arraystretch}{1.15}
\caption{Lesion ROI gaze response and artifact-related gaze attraction and attention cost by dataset}
\label{tab:results}
\newsavebox{\resultstabbox}
\sbox{\resultstabbox}{%
\begin{tabular}{@{}lccc@{}}
\toprule
 & \multicolumn{2}{c}{Retrospective paired-video} & Prospective real-time \\
\cmidrule(lr){2-3}\cmidrule(l){4-4}
Metric & Unassisted\textsuperscript{¶} & CADe-assisted & CADe-assisted\textsuperscript{§} \\
\midrule
\multicolumn{4}{@{}l}{\textit{Lesion ROI gaze response}} \\
Lesion ROI gaze recognition, proportion (n/N) & 0.980 (201/205) & 0.990 (203/205) & 0.867 (52/60) \\
First ROI-entry time, ms\textsuperscript{*} & 1458.3 & 1310.5 & 1633.5\textsuperscript{†} \\
\quad Senior & 1542.4 & 1246.7 & --- \\
\quad Novice & 1332.1 & 1406.1 & --- \\
\addlinespace
\multicolumn{4}{@{}l}{\textit{Artifact ROI gaze attraction}} \\
Attraction-triggered proportion (n/N) & --- & 0.486 (68/140) & 0.652 (533/817) \\
\quad Senior & --- & 0.524 (44/84) & --- \\
\quad Novice & --- & 0.429 (24/56) & --- \\
\addlinespace
\multicolumn{4}{@{}l}{\textit{Artifact attention cost}} \\
Artifact on-screen duration, median (s) & --- & 0.033 & 0.267 \\
Attention occupation duration, median (s)\textsuperscript{‡} & --- & 1.00 & 1.10 \\
Time amplification, median (ratio)\textsuperscript{‡} & --- & 17.55 & 5.15 \\
\quad Senior & --- & 23.26 & --- \\
\quad Novice & --- & 14.05 & --- \\
\bottomrule
\end{tabular}}
\usebox{\resultstabbox}\par\vspace{2pt}
\begin{minipage}{\wd\resultstabbox}
\scriptsize
\textit{Notes:} Retrospective lesion analyses included 5 endoscopists (3 senior, 2 novice) and 41 lesion events, yielding 205 subject--lesion observations per condition; artifact analyses included 28 events, yielding 140 subject--event observations per condition. The prospective dataset included 42 procedures, 60 lesion events, and 817 artifact events. The two datasets were analyzed separately and were not pooled. \textsuperscript{*} Means of participant-level summaries (all, n=5; senior, n=3; novice, n=2). \textsuperscript{†} Median among 52 recognition-positive lesion events. \textsuperscript{‡} Calculated among attraction-triggered events with complete recovery (retrospective n=57: senior 36, novice 21; prospective n=461); right-censored events were excluded. \textsuperscript{¶} Artifact-related metrics were not estimated for the unassisted condition under the study design. \textsuperscript{§} All prospective procedures were performed by senior endoscopists; no novice subgroup was available. ---, not estimated. CADe, computer-aided detection; ROI, region of interest.
\end{minipage}
\end{table}
\FloatBarrier

\subsection{Lesion-directed gaze recognition and first-entry time}\label{subsec:lesion-roi-gaze-recognition-and-first-entry-timing-under-normal-control-and-cade-assistance}

\begin{figure}[!htbp]
\centering
\includegraphics[width=\textwidth]{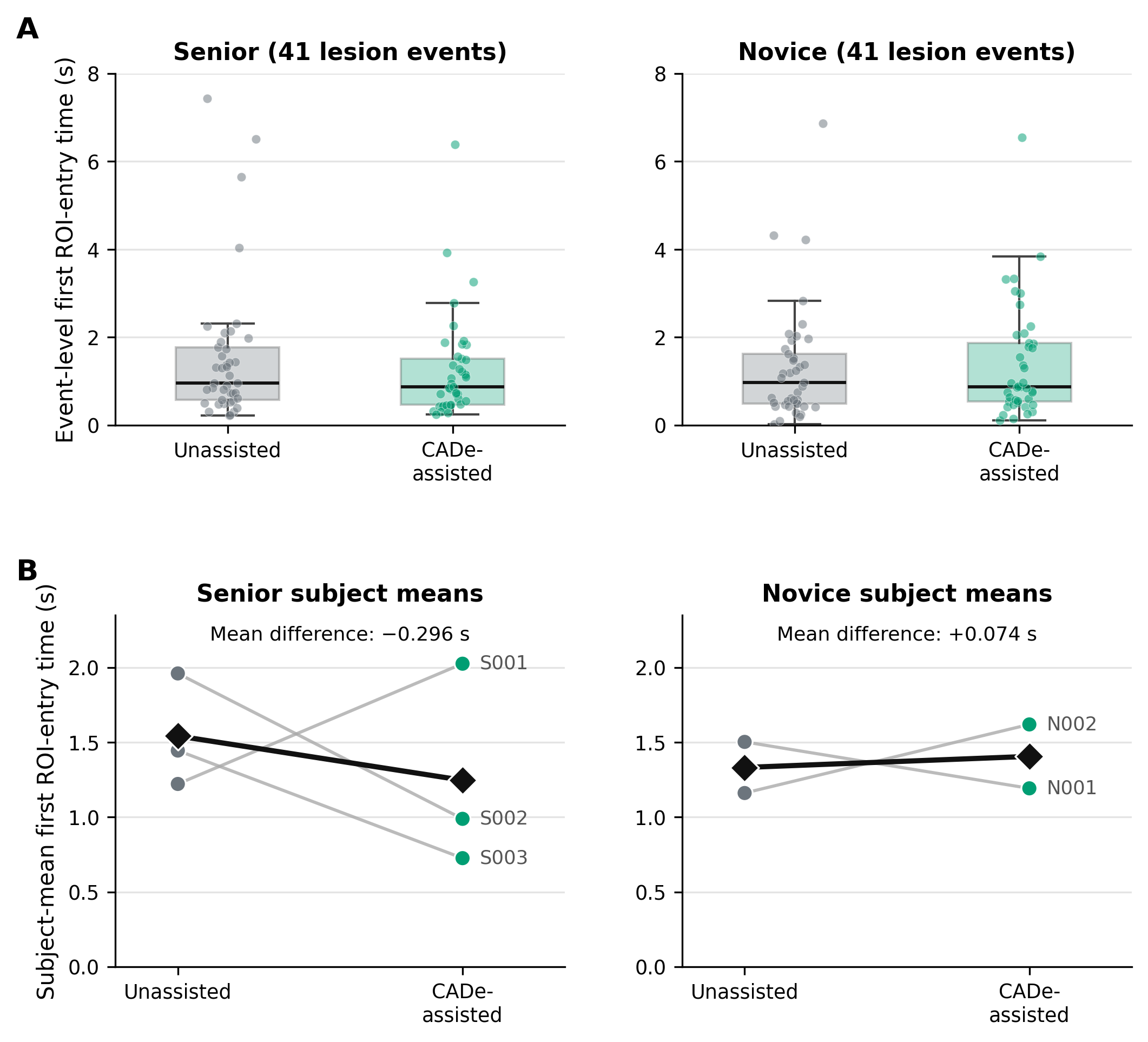}
\caption{First lesion ROI-entry timing under unassisted and CADe-assisted viewing. (A) Event-level first lesion ROI-entry time in senior and novice endoscopists during retrospective unassisted and CADe-assisted viewing. Each point represents a lesion-event estimate after averaging across subjects within the same group and condition, and boxplots summarize the event-level distributions. (B) Subject-mean first lesion ROI-entry time for the same retrospective paired subjects, with paired lines linking each subject's unassisted and CADe-assisted values; black diamonds indicate group means. Subject labels are anonymized sequentially within experience group (S, senior; N, novice). The senior subgroup showed earlier mean first ROI entry under CADe-assisted viewing, whereas the novice subgroup did not show the same direction of change. Figure 3 is interpreted as exploratory lesion-side gaze-mechanism evidence and should be read together with Table 2, which reports lesion ROI recognition rates and first-entry summaries.}
\label{fig:lesion}
\end{figure}

In the retrospective paired experiment, lesion gaze recognition was high under both unassisted and CADe-assisted conditions, at 98.0\% (201/205) and 99.0\% (203/205), respectively. Mean first ROI-entry time across the 5 participants was 1458.3 ms under the unassisted condition and 1310.5 ms with CADe assistance, corresponding to an earlier entry of 147.8 ms with CADe (Figure~\ref{fig:lesion}A,B). In the prospective real-time recordings, lesion gaze recognition was 86.7\% (52/60). Among recognition-positive lesion events, the median first ROI-entry time was 1633.5 ms (n=52) (Table~\ref{tab:results}).

\FloatBarrier

\subsection{Experience-stratified gaze responses}\label{subsec:exploratory-experience-stratified-analysis-of-lesion-and-artifact-gaze-metrics}

In the retrospective paired experiment, mean first lesion ROI-entry time was 295.6 ms shorter with CADe assistance among senior endoscopists (1542.4 vs 1246.7 ms), whereas it was 74.0 ms longer among novice endoscopists (1332.1 vs 1406.1 ms) (Figure~\ref{fig:lesion}A,B). For false-positive prompts, senior endoscopists showed a higher attraction-trigger rate than novice endoscopists (52.4\%, 44/84 vs 42.9\%, 24/56) and greater median time amplification (23.26-fold, n=36 vs 14.05-fold, n=21). These experience-stratified findings should be considered exploratory given the small subgroup sample (3 senior and 2 novice endoscopists).

\section{Discussion}\label{sec:discussion}

This study provides event-level eye-tracking evidence that false-positive CADe prompts can capture and sustain endoscopists' visual attention during colonoscopy. False-positive prompts attracted gaze in both the controlled paired-video experiment and prospective real-time recordings, and the associated attention occupation persisted for approximately 1000--1100 ms, substantially longer than the prompts themselves. As a secondary comparison, lesion gaze recognition remained high with and without CADe assistance, while first gaze entry into lesion ROIs occurred earlier with CADe. Exploratory analyses further suggested that prompt-related gaze responses may differ according to endoscopist experience. Together, these findings characterize how individual CADe prompts influence visual attention beyond conventional procedure-level detection outcomes.

The principal finding extends the evaluation of false-positive CADe burden from prompt frequency to prompt-level attentional dynamics. Previous studies have shown that CADe systems can generate multiple false-positive alerts during colonoscopy and that a greater false-positive burden may reduce the benefit of CADe assistance \citep{hassan_false_positive_classification,zhang_false_positive_assistance}. However, event counts indicate how often false-positive prompts occur but not how strongly an individual prompt affects visual search or how long that effect persists. In our study, false-positive prompts were followed by gaze attraction and attention occupation lasting approximately 1 s in both datasets. This indicates that even a brief visual prompt can be associated with a gaze response that persists well beyond its on-screen duration. Attention occupation duration therefore provides a time-based measure of prompt-related gaze persistence that complements conventional false-positive event rates.

Time amplification further quantified the persistence of this gaze response relative to prompt visibility. Median time amplification was 17.55-fold in the retrospective experiment and 5.15-fold in the prospective recordings. The difference between these values should be interpreted in the context of the substantially different prompt durations in the two datasets, with median durations of 33 ms retrospectively and 267 ms prospectively. Time amplification is therefore sensitive to the duration of the visual stimulus and is best interpreted together with absolute occupation time. We regard attention occupation duration as the primary temporal measure, while time amplification provides a complementary measure of how long the prompt-related gaze response persists relative to the prompt itself.

The lesion-directed analysis provides a useful comparison with the false-positive findings. Lesion gaze recognition was already high under unassisted viewing and increased only slightly with CADe assistance, whereas first gaze entry into the lesion ROI occurred 147.8 ms earlier with CADe. This suggests that CADe may primarily facilitate earlier gaze orientation toward visible lesions when baseline gaze recognition is already high. Eye-tracking studies have distinguished failure to fixate a lesion from failure to recognize it despite fixation \citep{ahmad_visual_recognition_errors}. The result should therefore not be interpreted as earlier lesion detection or improved diagnostic performance, which were not directly measured. Although the two sessions were separated by two weeks, the viewing order was fixed, and residual familiarity with the prerecorded videos cannot be excluded as a contributor to the observed first-entry difference.

Experience-stratified analyses suggested different gaze responses between senior and novice endoscopists. Senior endoscopists showed a larger CADe-associated reduction in lesion first-entry time, a higher attraction-trigger rate for false-positive prompts, and greater time amplification. Previous eye-tracking studies have similarly shown that visual-search patterns vary according to endoscopy experience and proficiency \citep{he_simulated_colonoscopy_gaze,karamchandani_trainee_gaze_patterns}. These findings may reflect differences in visual monitoring, search efficiency, or sensitivity to on-screen prompts rather than greater dependence on CADe. Given the small subgroup sample of 3 senior and 2 novice endoscopists, these results should be considered exploratory.

Prompt-level gaze responses also provide a potential mechanistic perspective on broader questions of human--AI collaboration. Budzyń et al.~\citep{budzyn_endoscopist_deskilling} reported lower unassisted ADR after repeated AI exposure, whereas Pedersen et al.~\citep{pedersen_learning_deskilling} found no persistent upskilling or deskilling after CADe withdrawal. Our study does not address whether repeated CADe exposure changes independent detection performance. Instead, it identifies a proximal behavioral response that occurs while CADe is being used. Future longitudinal studies could examine whether repeated prompt-related attentional capture accumulates over time and whether such exposure is associated with subsequent changes in unassisted visual search or lesion detection.

These findings may also inform the evaluation and design of future CADe systems. Current evaluation primarily considers clinical detection endpoints and the frequency of false-positive alerts. Reducing false-positive alerts to avoid alert fatigue has been identified as a research priority for AI-assisted colonoscopy \citep{ahmad_ai_colonoscopy_delphi}. However, systems with similar false-positive event rates may differ in the amount of visual attention consumed by each prompt. Prompt frequency and prompt-related attention should therefore be considered as complementary dimensions of CADe performance. Event-level measures such as attention occupation duration may help evaluate how prompt salience, persistence, and spatial presentation influence the endoscopist's visual search. Reducing unnecessary attentional capture while preserving rapid orientation toward true lesions may provide an additional design objective for more effective human--AI collaboration.

The retrospective and prospective datasets provide complementary evidence. The paired-video experiment enabled within-subject comparison of gaze responses to the same visual regions with and without visible CADe prompts, whereas the prospective recordings demonstrated that false-positive prompt-related gaze attraction and persistence were also observed during real-time clinical workflow. The two datasets differed in experimental setting, prompt frequency, and prompt duration and were therefore analyzed separately.

Several limitations should be considered. The retrospective experiment included only 5 endoscopists, limiting the precision and generalizability of the experience-stratified findings. The viewing order was not counterbalanced. The prospective dataset contained only CADe-assisted procedures and therefore provided real-time observational evidence without a matched prospective unassisted comparison. False-positive event identification depended on expert-defined lesion windows and the event-extraction pipeline, and event misclassification cannot be completely excluded. Gaze was recorded at 30 Hz, corresponding to a temporal resolution of approximately 33 ms, which limits the precision of very short temporal measurements \citep{andersson_sampling_frequency_eye_tracking} and makes time amplification particularly sensitive when prompt duration approaches a single video frame. Finally, the study evaluated one CADe system under specific display and eye-tracking conditions, and whether prompt-related attention occupation translates into measurable changes in lesion detection remains to be determined.
\section{Conclusions}\label{sec:conclusion}

False-positive CADe prompts frequently attracted endoscopists' gaze, and the associated attention occupation persisted substantially beyond prompt visibility in both controlled and real-time clinical settings. Event-level gaze metrics may complement conventional detection endpoints and false-positive counts by capturing how individual prompts shape visual attention. Incorporating prompt-related attentional effects into future CADe evaluation and design may support more effective human--AI collaboration.

\section*{ACKNOWLEDGMENT}
This work was supported by the Excellent Young Scientists Fund of Fujian Provincial Natural Science Foundation (Grant No. 2026D017).

\EndoGazeBibliography


@article{ahmad_ai_colonoscopy_delphi,
  author = {Ahmad, Omer F. and Mori, Yuichi and Misawa, Masashi and Kudo, Shin-ei and Anderson, John T. and Bernal, Jorge and Berzin, Tyler M. and Bisschops, Raf and Byrne, Michael F. and Chen, Peng-Jen and East, James E. and Eelbode, Tom and Elson, Daniel S. and Gurudu, Suryakanth R. and Histace, Aymeric and Karnes, William E. and Repici, Alessandro and Singh, Rajvinder and Valdastri, Pietro and Wallace, Michael B. and Wang, Pu and Stoyanov, Danail and Lovat, Laurence B.},
  title = {Establishing key research questions for the implementation of artificial intelligence in colonoscopy: a modified Delphi method},
  journal = {Endoscopy},
  year = {2021},
  volume = {53},
  number = {9},
  pages = {893--901},
  doi = {10.1055/a-1306-7590}
}

@article{ahmad_visual_recognition_errors,
  author = {Ahmad, Omer F. and Mazomenos, Evangelos and Chadebecq, Francois and Kader, Rawen and Hussein, Mohamed and Haidry, Rehan J. and Puyal, Juana Gonz{\'a}lez-Bueno and Brandao, Patrick and Toth, Daniel and Mountney, Peter and Seward, Ed and Vega, Roser and Stoyanov, Danail and Lovat, Laurence B.},
  title = {Identifying key mechanisms leading to visual recognition errors for missed colorectal polyps using eye-tracking technology},
  journal = {Journal of Gastroenterology and Hepatology},
  year = {2023},
  volume = {38},
  number = {5},
  pages = {768--774},
  doi = {10.1111/jgh.16127}
}

@article{almansa_visual_gaze_adenoma,
  author = {Almansa, Cristina and Shahid, Muhammad W. and Heckman, Michael G. and Preissler, Susan and Wallace, Michael B.},
  title = {Association Between Visual Gaze Patterns and Adenoma Detection Rate During Colonoscopy: A Preliminary Investigation},
  journal = {American Journal of Gastroenterology},
  year = {2011},
  volume = {106},
  number = {6},
  pages = {1070--1074},
  doi = {10.1038/ajg.2011.26}
}

@article{andersson_sampling_frequency_eye_tracking,
  author = {Andersson, Richard and Nystr{\"o}m, Marcus and Holmqvist, Kenneth},
  title = {Sampling frequency and eye-tracking measures: how speed affects durations, latencies, and more},
  journal = {Journal of Eye Movement Research},
  year = {2010},
  volume = {3},
  number = {3},
  note = {Article 6},
  doi = {10.16910/jemr.3.3.6}
}

@article{budzyn_endoscopist_deskilling,
  author = {Budzyń, Krzysztof and Romańczyk, Marcin and Kitala, Diana and Kołodziej, Paweł and Bugajski, Marek and Adami, Hans O. and Blom, Johannes and Buszkiewicz, Marek and Halvorsen, Natalie and Hassan, Cesare and Romańczyk, Tomasz and Holme, Øyvind and Jarus, Krzysztof and Fielding, Shona and Kunar, Melina and Pellise, Maria and Pilonis, Nastazja and Kamiński, Michał Filip and Kalager, Mette and Bretthauer, Michael and Mori, Yuichi},
  title = {Endoscopist deskilling risk after exposure to artificial intelligence in colonoscopy: a multicentre, observational study},
  journal = {The Lancet Gastroenterology \& Hepatology},
  year = {2025},
  volume = {10},
  number = {10},
  pages = {896--903},
  doi = {10.1016/S2468-1253(25)00133-5}
}

@article{chung_cade_false_positive_comparison,
  author = {Chung, Goh Eun and Lee, Jooyoung and Lim, Seon Hee and Kang, Hae Yeon and Kim, Jung and Song, Ji Hyun and Yang, Sun Young and Choi, Ji Min and Seo, Ji Yeon and Bae, Jung Ho},
  title = {A prospective comparison of two computer aided detection systems with different false positive rates in colonoscopy},
  journal = {npj Digital Medicine},
  year = {2024},
  volume = {7},
  number = {1},
  pages = {366},
  doi = {10.1038/s41746-024-01334-y}
}

@article{davila_pinon_computer_assisted_colonoscopy,
  author = {Davila-Piñón, Pedro and Díez-Martín, Astrid I. and Nogueira-Rodríguez, Alba and Fdez-Riverola, Florentino and Glez-Peña, Daniel and Reboiro-Jato, Miguel and De Castro, Luisa and Fernández-De Castro, Daniel and Vega, Pablo and Galovart-Araguas, Miguel Telmo and Soto, Santiago and Alonso-Lorenzo, Sara and Martínez-Turnes, Alfonso and Pin, Noel and Zarraquiños, Sara and Remedios, David and Sánchez-Gomez, Cristina and Souto-Rodríguez, Raquel and Baiocchi, Franco and Iglesias-Varela, María José and Ledo, Alejandro and Tejido-Sandoval, Coral and Rivas, Laura and García-Morales, Natalia and Rodríguez-De Jesus, Antonio and Puga, Manuel and Castiñeira-Domínguez, María Belén and Fernández-Fernández, Nereida and Germade-Martínez, Arantza and Gonzalez-De La Ballina, Enrique and Pérez-Medrano, Indhira and López-Fernández, Hugo and Cubiella, Joaquín},
  title = {Computer-assisted versus standard colonoscopy for adenoma detection in a population-based colorectal cancer screening program: a randomized clinical trial},
  journal = {Endoscopy},
  year = {2026},
  note = {Online ahead of print},
  doi = {10.1055/a-2895-1613}
}

@article{hassan_ai_validation,
  author = {Hassan, Cesare and Wallace, Michael B. and Sharma, Prateek and Maselli, Roberta and Craviotto, Vincenzo and Spadaccini, Marco and Repici, Alessandro},
  title = {New artificial intelligence system: first validation study versus experienced endoscopists for colorectal polyp detection},
  journal = {Gut},
  year = {2020},
  volume = {69},
  number = {5},
  pages = {799--800},
  doi = {10.1136/gutjnl-2019-319914}
}

@article{hassan_false_positive_classification,
  author = {Hassan, Cesare and Badalamenti, Matteo and Maselli, Roberta and Correale, Loredana and Iannone, Andrea and Radaelli, Franco and Rondonotti, Emanuele and Ferrara, Elisa and Spadaccini, Marco and Alkandari, Asma and Fugazza, Alessandro and Anderloni, Andrea and Galtieri, Piera Alessia and Pellegatta, Gaia and Carrara, Silvia and Di Leo, Milena and Craviotto, Vincenzo and Lamonaca, Laura and Lorenzetti, Roberto and Andrealli, Alida and Antonelli, Giulio and Wallace, Michael and Sharma, Prateek and Rösch, Thomas and Repici, Alessandro},
  title = {Computer-aided detection-assisted colonoscopy: classification and relevance of false positives},
  journal = {Gastrointestinal Endoscopy},
  year = {2020},
  volume = {92},
  number = {4},
  pages = {900--904.e4},
  doi = {10.1016/j.gie.2020.06.021}
}

@article{hassan_real_time_cade_meta_analysis,
  author = {Hassan, Cesare and Spadaccini, Marco and Mori, Yuichi and Foroutan, Farid and Facciorusso, Antonio and Gkolfakis, Paraskevas and Tziatzios, Georgios and Triantafyllou, Konstantinos and Antonelli, Giulio and Khalaf, Kareem and Rizkala, Tommy and Vandvik, Per Olav and Fugazza, Alessandro and Rondonotti, Emanuele and Glissen-Brown, Jeremy R. and Kamba, Shunsuke and Maida, Marcello and Correale, Loredana and Bhandari, Pradeep and Jover, Rodrigo and Sharma, Prateek and Rex, Douglas K. and Repici, Alessandro},
  title = {Real-Time Computer-Aided Detection of Colorectal Neoplasia During Colonoscopy: A Systematic Review and Meta-analysis},
  journal = {Annals of Internal Medicine},
  year = {2023},
  volume = {176},
  number = {9},
  pages = {1209--1220},
  doi = {10.7326/M22-3678}
}

@article{he_simulated_colonoscopy_gaze,
  author = {He, Wenjing and Bryns, Simon and Kroeker, Karen and Basu, Anup and Birch, Daniel and Zheng, Bin},
  title = {Eye gaze of endoscopists during simulated colonoscopy},
  journal = {Journal of Robotic Surgery},
  year = {2020},
  volume = {14},
  number = {1},
  pages = {137--143},
  doi = {10.1007/s11701-019-00950-1}
}

@article{holzwanger_false_positive_definitions,
  author = {Holzwanger, Erik A. and Bilal, Mohammad and Glissen Brown, Jeremy R. and Singh, Shailendra and Becq, Aymeric and Ernest-Suarez, Kenneth and Berzin, Tyler M.},
  title = {Benchmarking definitions of false-positive alerts during computer-aided polyp detection in colonoscopy},
  journal = {Endoscopy},
  year = {2021},
  volume = {53},
  number = {9},
  pages = {937--940},
  doi = {10.1055/a-1302-2942}
}

@article{ishibashi_ai_site_recognition,
  author = {Ishibashi, Fumiaki and Suzuki, Sho and Mochida, Kentaro and Nagai, Mizuki and Ozaki, Eri and Okusa, Kosuke},
  title = {Eye Tracking Analysis to Determine the Endoscopist's Recognition Rate for Artificial Intelligence-Detected Sites in Colonoscopy},
  journal = {Digestive Diseases and Sciences},
  year = {2025},
  volume = {70},
  number = {12},
  pages = {4113--4121},
  doi = {10.1007/s10620-025-09205-6}
}

@article{ito_cade_gaze_shift,
  author = {Ito, Shun and Ishibashi, Fumiaki and Okusa, Kosuke and Mochida, Kentaro and Tonishi, Takao and Ozaki, Eri and Suzuki, Sho},
  title = {Impact of Computer-Aided Detection on Endoscopist's Gaze-Shift Distance During Colonoscopy: a Randomized Controlled Trial (With Video)},
  journal = {Journal of Gastroenterology and Hepatology},
  year = {2026},
  volume = {41},
  number = {6},
  pages = {1760--1767},
  doi = {10.1111/jgh.70351}
}

@article{karamchandani_trainee_gaze_patterns,
  author = {Karamchandani, Urvi and Erridge, Simon and Evans-Harvey, Keane and Darzi, Ara and Hoare, Jonathan and Sodergren, Mikael Hans},
  title = {Visual gaze patterns in trainee endoscopists -- a novel assessment tool},
  journal = {Scandinavian Journal of Gastroenterology},
  year = {2022},
  volume = {57},
  number = {9},
  pages = {1138--1146},
  doi = {10.1080/00365521.2022.2064723}
}

@article{ladabaum_pragmatic_cade_trial,
  author = {Ladabaum, Uri and Shepard, John and Weng, Yingjie and Desai, Manisha and Singer, Sara J. and Mannalithara, Ajitha},
  title = {Computer-aided Detection of Polyps Does Not Improve Colonoscopist Performance in a Pragmatic Implementation Trial},
  journal = {Gastroenterology},
  year = {2023},
  volume = {164},
  number = {3},
  pages = {481--483.e6},
  doi = {10.1053/j.gastro.2022.12.004}
}

@article{lami_gaze_patterns_polyp_detection,
  author = {Lami, Mariam and Singh, Harsimrat and Dilley, James and Ashraf, Hajra and Edmondon, Matthew and Orihuela-Espina, Felipe and Hoare, Jonathan and Darzi, Ara and Sodergren, Mikael},
  title = {Gaze patterns hold key to unlocking successful search strategies and increasing polyp detection rate in colonoscopy},
  journal = {Endoscopy},
  year = {2018},
  volume = {50},
  number = {7},
  pages = {701--707},
  doi = {10.1055/s-0044-101026}
}

@article{nagai_optimal_visual_gaze,
  author = {Nagai, Mizuki and Ishibashi, Fumiaki and Okusa, Kosuke and Mochida, Kentaro and Ozaki, Eri and Morishita, Tetsuo and Suzuki, Sho},
  title = {Optimal visual gaze pattern of endoscopists for improving adenoma detection during colonoscopy (with video)},
  journal = {Gastrointestinal Endoscopy},
  year = {2025},
  volume = {101},
  number = {3},
  pages = {639--646.e3},
  doi = {10.1016/j.gie.2024.09.028}
}

@article{okumura_false_positive_detection,
  author = {Okumura, Taishi and Imai, Kenichiro and Misawa, Masashi and Kudo, Shin-Ei and Hotta, Kinichi and Ito, Sayo and Kishida, Yoshihiro and Takada, Kazunori and Kawata, Noboru and Maeda, Yuki and Yoshida, Masao and Yamamoto, Yoichi and Minamide, Tatsunori and Ishiwatari, Hirotoshi and Sato, Junya and Matsubayashi, Hiroyuki and Ono, Hiroyuki},
  title = {Evaluating false-positive detection in a computer-aided detection system for colonoscopy},
  journal = {Journal of Gastroenterology and Hepatology},
  year = {2024},
  volume = {39},
  number = {5},
  pages = {927--934},
  doi = {10.1111/jgh.16491}
}

@article{pedersen_learning_deskilling,
  author = {Pedersen, Tom Andre and Mori, Yuichi and Botteri, Edoardo and Engjom, Trond and Seip, Birgitte and Dimcevski, Georg Gjorgji and Havre, Roald Flesland},
  title = {Learning and deskilling effects of artificial intelligence in colonoscopy among endoscopists with different levels of experience: a pragmatic, prospective trial},
  journal = {Endoscopy},
  year = {2026},
  volume = {58},
  number = {9},
  pages = {1003--1014},
  doi = {10.1055/a-2858-7084}
}

@article{theeuwes_oculomotor_capture,
  author = {Theeuwes, Jan and Kramer, Arthur F. and Hahn, Sowon and Irwin, David E.},
  title = {Our Eyes Do Not Always Go Where We Want Them to Go: Capture of the Eyes by New Objects},
  journal = {Psychological Science},
  year = {1998},
  volume = {9},
  number = {5},
  pages = {379--385},
  doi = {10.1111/1467-9280.00071}
}

@article{tiankanon_false_positive_thresholds,
  author = {Tiankanon, Kasenee and Karuehardsuwan, Julalak and Aniwan, Satimai and Mekaroonkamol, Parit and Sunthornwechapong, Panukorn and Navadurong, Huttakan and Tantitanawat, Kittithat and Mekritthikrai, Krittaya and Samutrangsi, Salin and Vateekul, Peerapon and Rerknimitr, Rungsun},
  title = {Performance comparison between two computer-aided detection colonoscopy models by trainees using different false positive thresholds: a cross-sectional study in Thailand},
  journal = {Clinical Endoscopy},
  year = {2024},
  volume = {57},
  number = {2},
  pages = {217--225},
  doi = {10.5946/ce.2023.145}
}

@article{troya_cade_reaction_time_gaze,
  author = {Troya, Joel and Fitting, Daniel and Brand, Markus and Sudarevic, Boban and Kather, Jakob Nikolas and Meining, Alexander and Hann, Alexander},
  title = {The influence of computer-aided polyp detection systems on reaction time for polyp detection and eye gaze},
  journal = {Endoscopy},
  year = {2022},
  volume = {54},
  number = {10},
  pages = {1009--1014},
  doi = {10.1055/a-1770-7353}
}

@article{vanberkel_false_positive_continuous_interaction,
  author = {van Berkel, Niels and Opie, Jeremy and Ahmad, Omer F. and Lovat, Laurence and Stoyanov, Danail and Blandford, Ann},
  title = {Initial Responses to False Positives in {AI}-Supported Continuous Interactions: A Colonoscopy Case Study},
  journal = {ACM Transactions on Interactive Intelligent Systems},
  year = {2022},
  volume = {12},
  number = {1},
  pages = {1--18},
  doi = {10.1145/3480247}
}

@article{vasey_decide_ai,
  author = {Vasey, Baptiste and Nagendran, Myura and Campbell, Bruce and Clifton, David A. and Collins, Gary S. and Denaxas, Spiros and Denniston, Alastair K. and Faes, Livia and Geerts, Bart and Ibrahim, Mudathir and Liu, Xiaoxuan and Mateen, Bilal A. and Mathur, Piyush and McCradden, Melissa D. and Morgan, Lauren and Ordish, Johan and Rogers, Campbell and Saria, Suchi and Ting, Daniel S. W. and Watkinson, Peter and Weber, Wim and Wheatstone, Peter and McCulloch, Peter and {the DECIDE-AI expert group}},
  title = {Reporting guideline for the early-stage clinical evaluation of decision support systems driven by artificial intelligence: {DECIDE-AI}},
  journal = {Nature Medicine},
  year = {2022},
  volume = {28},
  number = {5},
  pages = {924--933},
  doi = {10.1038/s41591-022-01772-9}
}

@article{wang_real_time_polyp_detection,
  author = {Wang, Pu and Berzin, Tyler M. and Glissen Brown, Jeremy Romek and Bharadwaj, Shishira and Becq, Aymeric and Xiao, Xun and Liu, Peixi and Li, Liangping and Song, Yan and Zhang, Di and Li, Yi and Xu, Guangre and Tu, Mengtian and Liu, Xiaogang},
  title = {Real-time automatic detection system increases colonoscopic polyp and adenoma detection rates: a prospective randomised controlled study},
  journal = {Gut},
  year = {2019},
  volume = {68},
  number = {10},
  pages = {1813--1819},
  doi = {10.1136/gutjnl-2018-317500}
}

@article{zhang_false_positive_assistance,
  author = {Zhang, Chenxia and Yao, Liwen and Jiang, Ruiqing and Wang, Jing and Wu, Huiling and Li, Xun and Wu, Zhifeng and Luo, Renquan and Luo, Chaijie and Tan, Xia and Wang, Wen and Xiao, Bing and Hu, Huiyan and Yu, Honggang},
  title = {Assessment of the role of false-positive alerts in computer-aided polyp detection for assistance capabilities},
  journal = {Journal of Gastroenterology and Hepatology},
  year = {2024},
  volume = {39},
  number = {8},
  pages = {1623--1635},
  doi = {10.1111/jgh.16615}
}

@article{zhu_endoscopist_ai_gaze_dataset,
  author = {Zhu, Yan and Yang, Rui-Jie and Fu, Pei-Yao and Zhang, Zhen and Zhang, Yi-Zhe and Li, Quan-Lin and Wang, Shuo and Zhou, Ping-Hong},
  title = {Eye-tracking dataset of endoscopist-AI teaming during colonoscopy: Retrospective and real-time acquisition},
  journal = {Scientific Data},
  year = {2025},
  volume = {12},
  number = {1},
  pages = {212},
  doi = {10.1038/s41597-025-04535-6}
}
\end{document}